\documentclass[runningheads]{llncs}
\usepackage{graphicx}
\usepackage{amsmath,amssymb}
\usepackage{color}
\usepackage{bm}
\usepackage{breqn}

\newcommand{\etal}{\textit{et al}.}
\newcommand{\ie}{\textit{i}.\textit{e}.}

\def\proposedd{SCAN}
\def\proposedsub{\proposedd~Stage-1~}
\def\proposed{\proposedd~}
\DeclareMathOperator*{\argmin}{arg\,min}

\begin{document}

\title{Unsupervised Image-to-Image Translation with Stacked Cycle-Consistent Adversarial Networks}
\titlerunning{Unsupervised Image-to-Image Translation with SCAN}

\authorrunning{M. Li et al.}

\author{Minjun Li$^{1,2}$, Haozhi Huang$^2$, Lin Ma$^2$, Wei Liu$^2$, Tong Zhang$^2$, Yu-Gang Jiang$^{1}$\thanks{The corresponding author.}
}
\institute{
Shanghai Key Lab of Intelligent Information Processing, School of Computer Science, Fudan University
\and
Tencent AI Lab \\
\email{me@minjun.li, \{huanghz08, forest.linma\}@gmail.com,\\ wl2223@columbia.edu, tongzhang@tongzhang-ml.org, ygj@fudan.edu.cn} \\
}

\maketitle
\begin{abstract}
Recent studies on unsupervised image-to-image translation have made a remarkable progress by training a pair of generative adversarial networks with a cycle-consistent loss. 
However, such unsupervised methods may generate inferior results when the image resolution is high or the two image domains are of significant appearance differences, such as the translations between semantic layouts and natural images in the Cityscapes dataset. 
In this paper, we propose novel Stacked Cycle-Consistent Adversarial Networks (SCANs) by decomposing a single translation into multi-stage transformations, which not only boost the image translation quality but also enable higher resolution image-to-image translations in a coarse-to-fine manner.
Moreover, to properly exploit the information from the previous stage, an adaptive fusion block is devised to learn a dynamic integration of the current stage's output and the previous stage's output.
Experiments on multiple datasets demonstrate that our proposed approach can improve the translation quality compared with previous single-stage unsupervised methods.


\end{abstract}

\section{Introduction}
Image-to-image translation attempts to convert the image appearance from one domain to another while preserving the intrinsic image content.
Many computer vision tasks can be formalized as a certain image-to-image translation problem, such as super-resolution~\cite{ledig2016photo,shi2016real}, image colorization~\cite{zhang2016colorful,zhang2017real,iizuka2016let}, image segmentation \cite{long2015fully,eigen2015predicting}, and image synthesis 
\cite{chen2017photographic,simo2016learning,xie2015holistically,laffont2014transient,bozhao2017arxiv}. 
However, conventional image-to-image translation methods are all task specific.
A common framework for universal image-to-image translation remains as an emerging research subject in the literature, which has gained considerable attention in recent studies
\cite{isola2016image,zhu2017unpaired,kim2017learning,liu2017unsupervised,yi2017dualgan}.

\begin{figure}[t]
\begin{center}
  \centering
  \resizebox{1\linewidth}{!}{
   \begin{minipage}[t]{1.0\linewidth}
   \centering
   \includegraphics[width=0.85\linewidth]{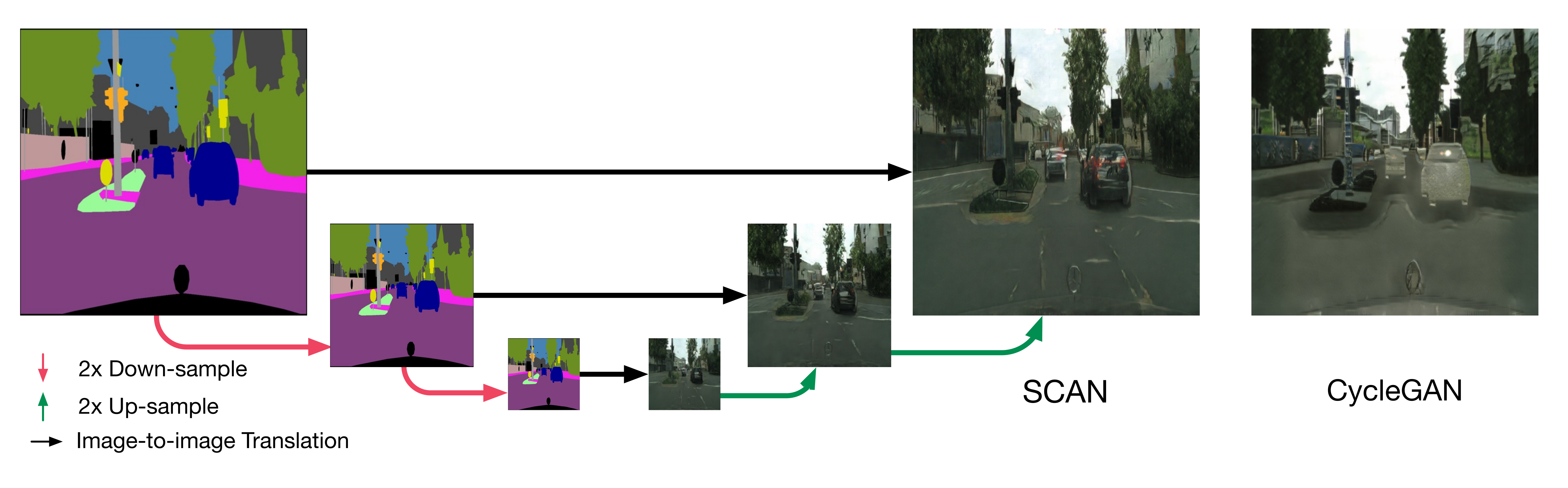}
   \end{minipage}
  }
  \caption{
  Given unpaired images from two domains, our proposed \proposed learns the image-to-image translation by a stacked structure in a coarse-to-fine manner.
  For the Cityscapes \textit{Labels $\rightarrow$ Photo} task in $512 \times 512$ resolution, the result of \proposed (left) appears more realistic and includes finer details compared with the result of CycleGAN~\cite{zhu2017unpaired} (right).}
  \label{fig:1}
\end{center}
\end{figure}

Isola \etal~\cite{isola2016image} leveraged the power of 
generative adversarial networks (GANs) \cite{goodfellow2014generative,mirza2014conditional,bozhao2018arxiv}, which encourage the translation results to be indistinguishable from real images in the target domain, 
to learn image-to-image translation from image pairs in a supervised fashion.
However, obtaining pairwise training data is time-consuming and heavily relies on human labor.
Recent works \cite{zhu2017unpaired,kim2017learning,liu2017unsupervised,yi2017dualgan} explore tackling the image-to-image translation problem without using pairwise data.
Under the unsupervised setting, besides the traditional adversarial loss used in supervised image-to-image translation, a cycle-consistent loss is introduced to restrain the two cross-domain transformations $G$ and $F$ to be the inverses of each other (\ie, $G(F(x))\approx x$ and $G(F(y))\approx y$).
By constraining both of the adversarial and cycle-consistent losses, the networks learn how to accomplish cross-domain transformations without using pairwise training data.

Despite the progress mentioned above, existing unsupervised image-to-image translation methods may generate inferior results when two image domains are of significant appearance differences or the image resolution is high.
As shown in Figure \ref{fig:1}, the result of CycleGAN~\cite{zhu2017unpaired} in translating a Cityscapes semantic layout to a realistic picture lacks details and remains visually unsatisfactory.
The reason for this phenomenon lies in the significant visual gap between the two distinct image domains, which makes the cross-domain transformation too complicated to be learned by running a single-stage unsupervised approach.

Jumping out of the scope of unsupervised image-to-image translation, many methods have leveraged the power of multi-stage refinements to tackle image generation from latent vectors~\cite{denton2015deep,karras2017pregressive}, caption-to-image~\cite{zhang2016stackgan}, and supervised image-to-image translation~\cite{chen2017photographic,eigen2015predicting,wang2017high}.
By generating an image in a coarse-to-fine manner, a complicated transformation is broken down into easy-to-solve pieces.
Wang \etal~\cite{wang2017high} successfully tackled the high-resolution image-to-image translation problem in such a coarse-to-fine manner with multi-scale discriminators.
However, their method relies on pairwise training images, so cannot be directly applied to our studied unsupervised image-to-image translation task.
To the best of our knowledge, there exists no attempt to exploit stacked networks to overcome the difficulties encountered in learning unsupervised image-to-image translation.

In this paper, we propose the stacked cycle-consistent adversarial networks (SCANs) aiming for unsupervised learning of image-to-image translation. 
We decompose a complex image translation into multi-stage transformations, including a coarse translation followed by multiple refinement processes.
The coarse translation learns to sketch a primary result in low-resolution.
The refinement processes improve the translation by adding details into the previous results to produce higher resolution outputs.
We adopt a conjunction of an adversarial loss and a cycle-consistent loss in all stages to learn translations from unpaired image data.
To benefit more from multi-stage learning, we also introduce an adaptive fusion block in the refinement processes to learn the dynamic integration of the current stage's output and the previous stage's output. 
Extensive experiments demonstrate that our proposed model can not only generate results with realistic details, but also enable us to learn unsupervised image-to-image translation in higher resolution.

In summary, our contributions are mainly two-fold. Firstly, we propose SCANs to model the unsupervised image-to-image translation problem in a coarse-to-fine manner for generating results with finer details in higher resolution.
Secondly, we introduce a novel adaptive fusion block to dynamically integrate the current stage's output and the previous stage's output, which outperforms directly stacking multiple stages.

\section{Related Work}

\smallskip
\noindent\textbf{Image-to-image translation.} GANs~\cite{goodfellow2014generative} have shown impressive results in a wide range of tasks including super-resolution \cite{ledig2016photo,shi2016real}, video generation \cite{xiong2018gan}, image colorization \cite{isola2016image}, image style transfer \cite{zhu2017unpaired} etc. 
The essential part of GANs is the idea of using an adversarial loss that encourages the translated results to be indistinguishable from real target images.
Among the existing image-to-image translation works using GANs, perhaps the most well-known one would be Pix2Pix \cite{isola2016image}, in which Isola \etal~applied GANs with a regression loss to learn pairwise image-to-image translation.
Due to the fact that pairwise image data is difficult to obtain, image-to-image translation using unpaired data has drawn rising attention in recent studies.
Recent works by Zhu \etal~\cite{zhu2017unpaired}, Yi \etal~\cite{yi2017dualgan}, and Kim \etal~\cite{kim2017learning} have tackled the image translation problem using a combination of adversarial and cycle-consistent losses. 
Taigman \etal~\cite{taigman2016unsupervised} applied cycle-consistency in the feature level with the adversarial loss to learn a one-side translation from unpaired images.
Liu \etal~\cite{liu2017unsupervised} used a GAN combined with Variational Auto Encoder (VAE) to learn a shared latent space of two given image domains. 
Liang \etal~\cite{liang2017generative} combined the ideas of adversarial and contrastive losses, using a contrastive GAN with cycle-consistency to learn the semantic transform of two given image domains with labels.
Instead of trying to translate one image to another domain directly, our proposed approach focuses on exploring refining processes in multiple steps to generate a more realistic output with finer details by harnessing unpaired image data.

\smallskip
\noindent\textbf{Multi-stage learning.}
Extensive works have proposed to choose multiple stages to tackle complex generation or transformation problems.
Eigen \etal~\cite{eigen2015predicting} proposed a multi-scale network to predict depth, surface, and segmentation, which learns to refine the prediction result from coarse to fine. 
S2GAN introduced by Wang \etal~\cite{wang2016generative} utilizes two networks arranged sequentially to first generate a structure image and then transform it into a natural scene.
Zhang \etal~\cite{zhang2016stackgan} proposed StackGAN to generate high-resolution images from texts, which consists of two stages: the Stage-I network generates a coarse, low-resolution result, while the Stage-II network refines the result into a high-resolution, realistic image.
Chen \etal~\cite{chen2017photographic} applied a stacked refinement network to generate scenes from segmentation layouts.
To accomplish generating high-resolution images from latent vectors, Kerras \etal~\cite{karras2017pregressive} started from generating a $4\times 4$ resolution output, and then progressively stacked up both a generator and a discriminator to generate a $1024 \times 1024$ realistic image.
Wang \etal~\cite{wang2017high} applied a coarse-to-fine generator with a multi-scale discriminator to tackle the supervised image-to-image translation problem.
Different form the existing works, this work exploits stacked image-to-image translation networks coupled with a novel adaptive fusion block to tackle the unsupervised image-to-image translation problem.

\section{Proposed Approach}

\subsection{Formulation}
Given two image domains $X$ and $Y$, the mutual translations between them can be denoted as two mappings $G:X \to Y$ and $F:Y \to X$, each of which takes an image from one domain and translates it to the corresponding representation in the other domain.
Existing unsupervised image-to-image translation approaches~\cite{zhu2017unpaired,yi2017dualgan,kim2017learning,liu2017unsupervised,taigman2016unsupervised} finish the learning of $G$ and $F$ in a single stage, which generate results lacking details and are unable to handle complex translations.

In this paper, we decompose translations $G$ and $F$ into multi-stage mappings. For simplicity, now we describe our method in a two-stage setting. Specifically, we decompose $G = G_{2} \circ G_{1}$ and $F = F_{2} \circ F_{1}$. $G_{1}$ and $F_{1}$ (\textbf{Stage-1}) perform the cross-domain translation in a coarse scale, while $G_{2}$ and $F_{2}$ (\textbf{Stage-2}) serve as refinements  on the top of the outputs from the previous stage.
We first finish the training of Stage-$1$ in low-resolution and then train Stage-$2$ to learn refinement in higher resolution based on the fixed Stage-$1$.

Training two stages in the same resolution would make Stage-$2$ difficult to bring further improvement, as Stage-$1$ has already been optimized with the same objective function (see Section \ref{sec:comp}). 
On the other hand, we find that learning in a lower resolution allows the model to generate visually more natural results, since the manifold underlying the low-resolution images is easier to model.
Therefore, first, we constrain Stage-$1$ to train on 2x down-sampled image samples, denoted by $X_{\downarrow} $ and $Y_{\downarrow}$, to learn a base transformation. Second, based on the outputs of Stage-$1$, we train Stage-$2$ with image samples $X$ and $Y$ in the original resolution.
Such a formulation exploits the preliminary low-resolution results of Stage-$1$ and guides Stage-$2$ to focus on up-sampling and adding finer details, thus helping improve the overall translation quality.

In summary, to learn cross-domain translations $G: X \to Y$ and $F: Y \to X$ on given domains $X$ and $Y$, we first learn preliminary translations $G_1: X_{\downarrow} \to Y_{\downarrow}$ and $F_1: Y_{\downarrow} \to X_{\downarrow}$ at the 2x down-sampled scale. Then we use $G_{2}: X_{\downarrow} \to X$ and $F_2: Y_{\downarrow} \to Y$ to obtain the final output with finer details in the original resolution. Notice that we can iteratively decompose $G_2$ and $F_2$ into more stages.%

\subsection{Stage-1: Basic Translation}
\label{sec:stage1}

\begin{figure}[t]
\begin{center}
   \includegraphics[width=0.4\linewidth]{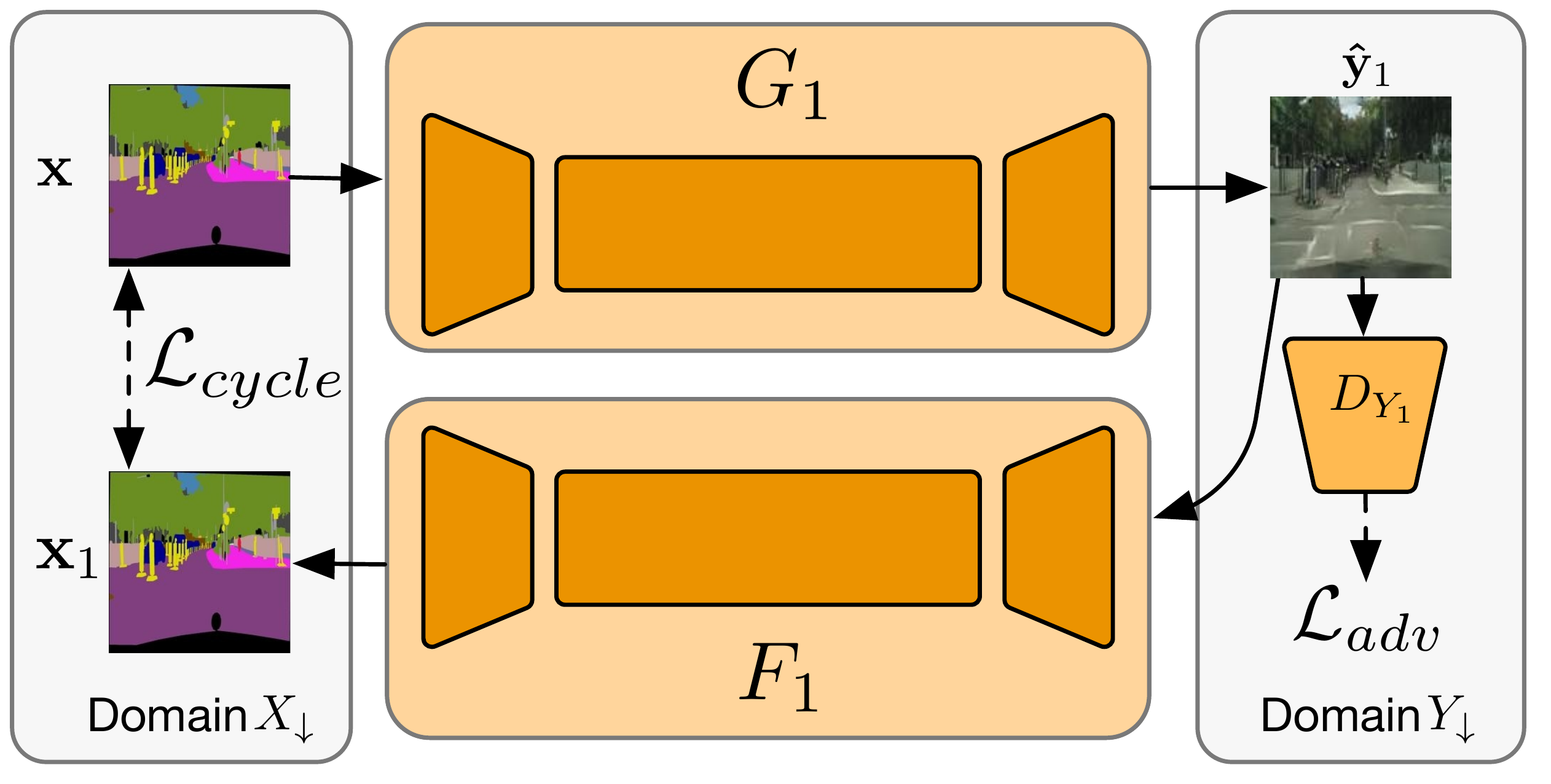}
\end{center}
   \caption{Illustration of an overview of Stage-$1$ for learning coarse translations in low-resolution under an unsupervised setting. Solid arrow denotes an input-output, and dashed arrow denotes a loss.}
\label{fig:stage1}
\end{figure}

In general, our Stage-$1$ module adopts a similar architecture of CycleGAN~\cite{zhu2017unpaired}, which consists of two image translation networks $G_{1}$, and $F_{1}$ and two discriminators $D_{X_1}, D_{Y_1}$. Note that Stage-$1$ is trained in low-resolution image domains $X_{\downarrow}$ and $Y_{\downarrow}$. 
Figure \ref{fig:stage1} shows an overview of the Stage-$1$ architecture.

Given a sample $\mathbf{x}_1 \in X_{\downarrow}$, $G_1$ translates it to a sample $\mathbf{\hat{y}}_1 = G_1(\mathbf{x}_1)$ in the other domain $Y_{\downarrow}$. On one hand, the discriminator $D_{Y_1}$ learns to classify the generated sample $\mathbf{\hat{y}}_1$ to class $0$ and the real image $\mathbf{y}$ to class $1$, respectively. On the other hand, $G_1$ learns to deceive $D_{Y_1}$ by generating more and more realistic samples. This can be formulated as an adversarial loss:
\begin{align}\begin{split}
\mathcal{L}_{adv}(G_1, D_{Y_1}, & X_\downarrow, Y_\downarrow) = 
\mathbb{E}_{\mathbf{y} \sim Y\downarrow}\left[\log(D_{Y_1}(\mathbf{y}))\right] \\
&+ \mathbb{E}_{\mathbf{x} \sim X\downarrow}\left[\log(1-D_{Y_1}(G_1(\mathbf{x})))\right].
\end{split}\end{align}
While $D_{Y_1}$ tries to maximize $\mathcal{L}_{adv}$, $G_1$ tries to minimize it. 
Afterward, we use $F_1$ to translate $\mathbf{\hat{y}}_1$ back to the domain $X_{\downarrow}$, and constrain $F_1(\mathbf{\hat{y}}_1=G_1(\mathbf{x}))$ to be close to the input $\mathbf{x}$. This can be formulated as a cycle-consistent loss:
\begin{equation}
\mathcal{L}_{cycle}(G_1, F_1, X_\downarrow) = \mathbb{E}_{\mathbf{x}\sim X\downarrow}\left\|\mathbf{x} - F_1(G_1(\mathbf{x}))\right\|_1.
\end{equation}

Similarly, for a sample $\mathbf{y}_1 \in Y_{\downarrow}$, we use $F_1$ to perform translation, use $D_{X_1}$ to calculate the adversarial loss, and then use $G_1$ to translate backward to calculate the cycle-consistent loss. Our full objective function for Stage-$1$ is a combination of the adversarial loss and the cycle-consistent loss:
\begin{dmath}
\mathcal{L}_{Stage1} =
\mathcal{L}_{adv}(G_1, D_{Y_1}, X_{\downarrow}, Y{\downarrow}) + 
\mathcal{L}_{adv}(F_1, D_{X_1}, Y_{\downarrow}, X_{\downarrow}) + 
\lambda[\mathcal{L}_{cycle}(G_1, F_1, X_{\downarrow}) +
\mathcal{L}_{cycle}(F_1, G_1, Y_{\downarrow})],
\label{equ:loss}
\end{dmath}
where $\lambda$ denotes the weight of the cycle-consistent loss.
We obtain the translations $G_1$ and $F_1$ by optimizing the following objective function: 
\begin{equation}
G_1, F_1 = \argmin_{G_1,F_1} \max_{D_{X_1}, D_{Y_1}} \mathcal{L}_{Stage1},
\label{equ:minmax}
\end{equation}
which encourages these translations to transform the results to another domain while preserving the intrinsic image content. 
As a result, the optimized translations $G_1$ and $F_1$ can perform a basic cross-domain translation in low resolution. 

\subsection{Stage-2: Refinement}
\label{sec:stage2}

\begin{figure*}[t]
\begin{center}
   \includegraphics[trim={0 0 0 0},clip,width=0.87\linewidth]{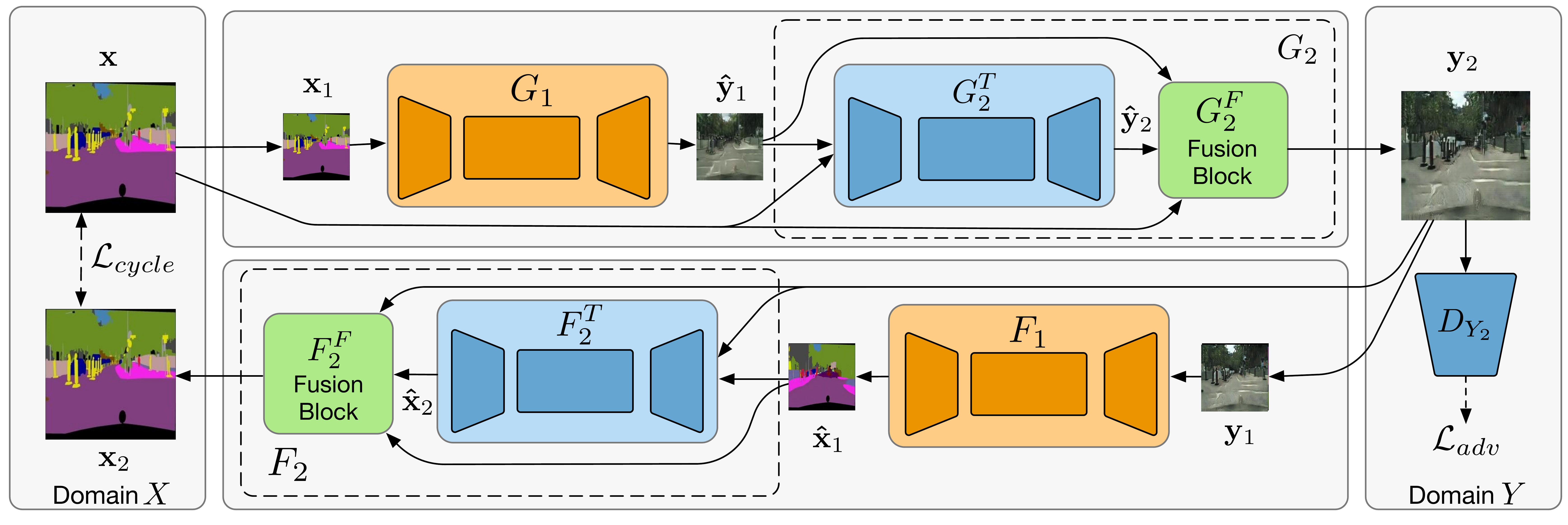}
\end{center}
   \caption{Illustration of an overview of our Stage-$2$ for learning refining processes on the top of Stage-$1$'s outputs. $G_{1}$ and $F_{1}$ are the translation networks learned in Stage-$1$. 
   In the training process, we keep the weights of $G_1$ and $F_1$ fixed. Solid arrow denotes an input-output, and dashed arrow denotes a loss.}
\label{fig:overview}
\end{figure*}

Since it is difficult to learn a complicated translation with the limited ability of a single stage, the translated output of Stage-$1$ may seem plausible but still leaves us much room for improvement.
To refine the output of Stage-$1$, we deploy Stage-$2$ with a stacked structure built on the top of the trained Stage-$1$ to complete the full translation to generate higher resolution results with finer details.

Stage-$2$ consists of two translation networks $G_2$, $F_2$ and two discriminator networks $D_{X_2}$, $D_{Y_2}$, as shown in Figure \ref{fig:overview}.
We only describe the architecture of $G_{2}$, since $F_{2}$ shares the same design (see Figure \ref{fig:overview}).

$G_2$ consists of two parts: a newly initialized image translation network $G_2^T$ and an adaptive fusion block $G_2^F$.
Given the output of Stage-$1$ ($\mathbf{\hat{y}}_1 = G_1(\mathbf{x}_1)$), we use nearest up-sampling to resize it to match the original resolution.
Different from the image translation network in Stage-$1$, which only takes $\mathbf{x} \in X$ as input, in Stage-$2$ we use both the current stage's input $\mathbf{x}$ and the previous stage's output $\mathbf{\hat{y}}_1$. Specifically, we concatenate $\mathbf{\hat{y}}_1$ and $\mathbf{x}$ along the channel dimension, and utilize $G_2^T$ to obtain the refined result
$\mathbf{\hat{y}}_2 = G_{2}^T(\mathbf{\hat{y}}_1, \mathbf{x})$.

\begin{figure}[t]
\begin{center}
   \includegraphics[width=0.65\linewidth]{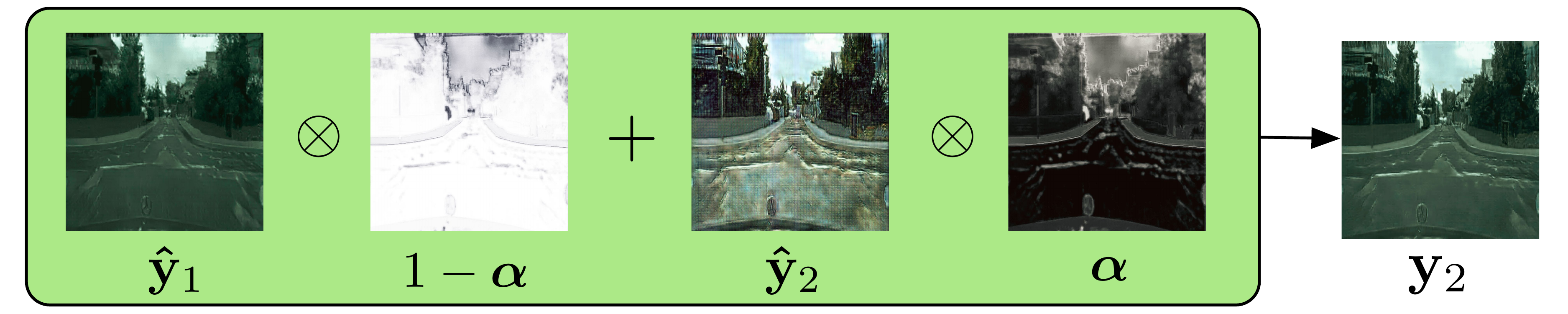}
\end{center}
   \caption{Illustration of the linear combination in an adaptive fusion block. The fusion block applies the fusion weight map $\bm\alpha$ to find defects in the previous result $\mathbf{\hat y}_1$ and correct it precisely using $\mathbf{\hat y}_2$ to produce a refined output $\mathbf{y}_2$.}
\label{fig:alpha}
\end{figure}

Besides simply using $\mathbf{\hat{y}}_2$ as the final output, we introduce an adaptive fusion block $G_2^F$ to learn a dynamic combination of $\mathbf{\hat{y}}_2$ and $\mathbf{\hat{y}}_1$ to fully utilize the entire two-stage structure. Specifically, the adaptive fusion block learns a pixel-wise linear combination of the previous results: 
\begin{equation}
G_2^F(\mathbf{\hat{y}}_1, \mathbf{\hat{y}}_2) = \mathbf{\hat{y}_1} \odot (1-\bm{\alpha}_{x}) + \mathbf{\hat{y}_2} \odot \bm{\alpha}_{x},
\label{equ:alpha}
\end{equation}
where $\odot$ denotes element-wise product and $\bm{\alpha} \in (0,1)^{H \times W}$ represents the fusion weight map, which is predicted by a convolutional network $h_{x}$:
\begin{equation}
\bm{\alpha}_{x} =  h_x(\mathbf{x}, \mathbf{\hat{y}}_1, \mathbf{\hat{y}}_2).
\label{equ:alphah}
\end{equation}
Figure \ref{fig:alpha} shows an example of adaptively combining the outputs from two stages. 

Similar to Stage-$1$, we use a combination of adversarial  and cycle-consistent losses to formulate our objective function of Stage-$2$:
\begin{dmath}
\mathcal{L}_{Stage2} = \mathcal{L}_{adv}(G_2 \circ G_1, D_{Y_2}, X, Y) + \mathcal{L}_{adv}(F_2 \circ F_1, D_{X_2}, Y, X) +  \lambda \left[\mathcal{L}_{cycle}(G_2 \circ G_1, F_2 \circ F_1, X) + \mathcal{L}_{cycle}(F_2 \circ F_1, G_2 \circ G_1, Y)\right].
\label{equ:loss2}
\end{dmath}
Optimizing this objective is similar to solving Equation \ref{equ:minmax}.
The translation networks $G_2$ and $F_2$ are learned to refine the previous results by correcting defects and adding details on them.

Finally, we complete our desired translations $G$ and $F$ by integrating the transformations in Stage-$1$ and Stage-$2$, which are capable of tackling a complex image-to-image translation problem under the unsupervised setting.

\section{Experiments}
The proposed approach is named \textit{SCAN} or \textit{SCAN Stage-$N$} if it has $N$ stages in the following experiments.
We explore several variants of our model to evaluate the effectiveness of our design in Section \ref{sec:ablation}. In all experiments, we decompose the target translation into two stages, except for exploring the ability of the three-stage architecture in high-resolution tasks in Section~\ref{sec:highres}.

We used the official released model of CycleGAN~\cite{zhu2017unpaired}
and Pix2Pix~\cite{isola2016image}
for $256\times256$ image translation comparisions.
For $512\times512$ tasks, we train the CycleGAN with the official code
since there is no available pre-trained model.

\subsection{Network Architecture}

For the image translation network, we follow the settings of \cite{zhu2017unpaired,liang2017generative}, adopting the encoder-decoder architecture from Johnson \etal~\cite{johnson2016perceptual}.
The network consists of two down-sample layers implemented by stride-2 convolution, six residual blocks and two up-sample layers implemented by sub-pixel convolution \cite{shi2016real}.
Note that different from \cite{zhu2017unpaired}, which used the fractionally strided convolution as the up-sample block, we use the sub-pixel convolution \cite{shi2016real}, for avoiding checkerboard artifacts \cite{odena2016deconvolution}.
The adaptive fusion block is a simple 3-layer convolutional network, which calculates the fusion weight map $\bm\alpha$ using two Convolution-InstanceNorm-ReLU blocks followed by a Convolution-Sigmoid block.
For the discriminator, we use the PatchGAN structure introduced in \cite{isola2016image}.

\subsection{Datasets}

To demonstrate the capability of our proposed method for tackling the complex image-to-image translation problem under unsupervised settings, we first conduct experiments on the Cityscapes dataset \cite{cordts2016cityscapes}. 
We compare with the state-of-the-art approaches in the \textit{Labels $\leftrightarrow$ Photo} task in $256 \times 256$ resolution.
To further show the effectiveness of our method to learn complex translations, we also extended the input size to a challenging $512 \times 512$ resolution, namely the high-resolution Cityscapes \textit{Labels $\rightarrow$ Photo} task.

Besides the \textit{Labels $\leftrightarrow$ Photo} task, we also select six image-to-image translation tasks from \cite{zhu2017unpaired}, including \textit{Map$\leftrightarrow$Aerial}, \textit{Facades$\leftrightarrow$Labels}
and \textit{Horse$\leftrightarrow$Zebra}. We compare our method with the CycleGAN \cite{zhu2017unpaired} in these tasks in $256 \times 256$ resolution.

\subsection{Training Details}

Networks in Stage-$1$ are trained from scratch, while networks in Stage-N are trained with the \{Stage-$1$, $\cdots$, Stage-(N-1)\}  networks fixed. For the GAN loss, Different from the previous works \cite{zhu2017unpaired,isola2016image}, we adopt a gradient penalty term $\lambda_{gp}(||\nabla D(x)||_2 - 1)^2$ in the discriminator loss to achieve a more stable training process~\cite{kodali2017train}. 
For all datasets, the Stage-$1$ networks are trained in $128 \times 128$ resolution, the Stage-$2$ networks are trained in $256 \times 256$ resolution. For the three-stage architecture in Section~\ref{sec:highres},  the Stage-$3$ networks are trained in $512 \times 512$ resolution.
We set batch size to $1$, $\lambda = 10$ and $\lambda_{\text{gp}} =10$ in all experiments.
All stages are trained with $100$ epochs for all datasets. We use Adam \cite{kingma2014adam} to optimize our networks with an initial learning rate as $0.0002$, and decrease it linearly to zero in the last $50$ epochs.

\subsection{Evaluation Metrics} 

\noindent\textbf{FCN Score and Segmentation Score.}
For the Cityscapes dataset, we adopt the FCN Score and the Segmentation Score as evaluation metrics from~\cite{isola2016image} for the \textit{Labels $\rightarrow$ Photo} task and the \textit{Photo $\rightarrow$ Labels} task, respectively.
The FCN Score employs an off-the-shelf FCN segmentation network \cite{long2015fully} to estimate the realism of the translated images.
The Segmentation Score includes three standard segmentation metrics, which are the per-pixel accuracy, the per-class accuracy, and the mean class accuracy, as defined in~\cite{long2015fully}.

\noindent\textbf{PSNR and SSIM.}
Besides using the FCN Score and the Segmentation Score, we also calculate the PSNR and the SSIM\cite{wang2004image} for a quantitative evaluation.
We apply the above metrics on the \textit{Map $\leftrightarrow$ Aerial} task and the \textit{Facades $\leftrightarrow$ Labels} task to measure both the color similarity and the structural similarity between the translated outputs and the ground truth images.

\noindent\textbf{User Preference.} \label{sec:userstudy}
We run user preference tests in the high-resolution Cityscapes \textit{Labels $\rightarrow$ Photos} task and the \textit{Horse$\rightarrow$Zebra} tasks
for evaluating the realism of our generated photos. 
In the user preference test, each time a user is presented with a pair of results from our proposed \proposed and the CycleGAN~\cite{zhu2017unpaired}, and asked which one is more realistic.
Each pair of the results is translated from the same image.
Images are all shown 
in randomized order.
In total, $30$ images from the Cityscapes test set and $10$ images from the Horse2Zebra test set are used in the user preference tests.
As a result, $20$ participates make a total of $600$ and $200$ preference choices, respectively.

\subsection{Comparisons} \label{sec:comp}

\begin{figure*}[t]
\begin{center}
   \includegraphics[width=\linewidth]{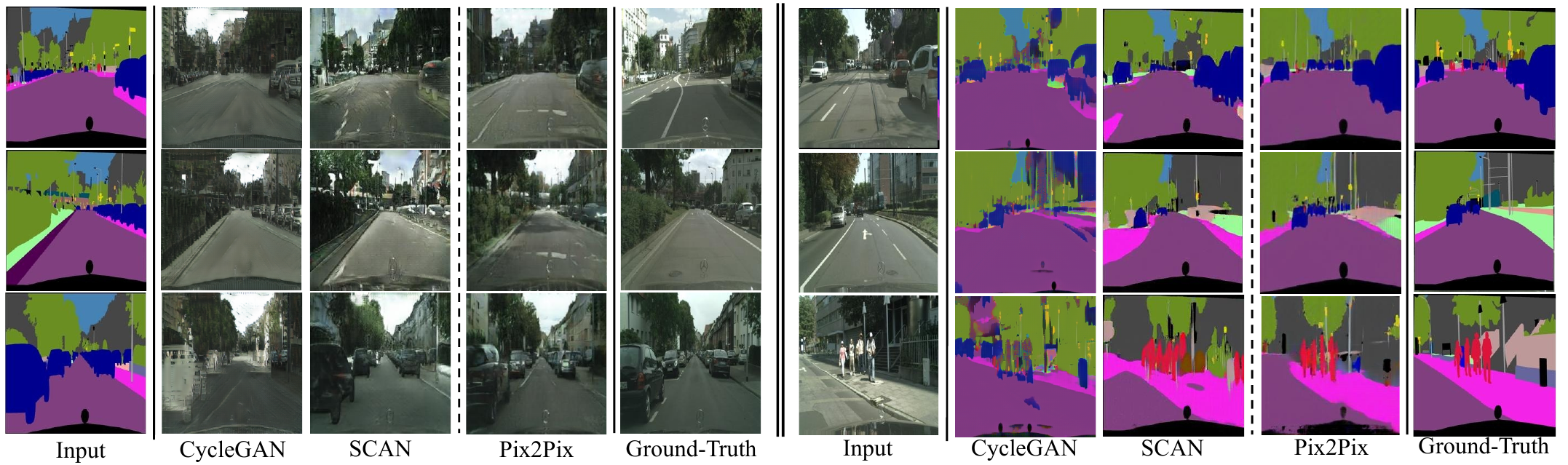}
\end{center}
   \caption{Comparisons on the Cityscapes dataset of $256\times256$ resolution. The left subfigure are \textit{Labels $\rightarrow$ Photo} results and the right are \textit{Photo $\rightarrow$ Labels} results. In the \textit{Labels $\rightarrow$ Photo} task, our proposed \proposed generates more natural photographs than CycleGAN; in the \textit{Photo $\rightarrow$ Labels} task, \proposed produces an accurate segmentation map while CycleGAN's results are blurry and suffer from deformation. SCAN also generates results that are visually closer to those of the supervised approach Pix2Pix than results of CycleGAN. Zoom in for better view.}
\label{fig:comp-city}
\end{figure*}

\begin{table}[t]
\small
\tabcolsep=0.11cm
\caption{FCN Scores in the Labels $\rightarrow$ Photo task and Segmentation Scores in the Photo $\rightarrow$ Labels task on the Cityscapes dataset. The proposed methods are named after \textit{SCAN (Stage-1 resolution)-(Stage-2 resolution)}. \textit{FT} means that we also \textit{fine-tune} the Stage-1 model instead of fixing its weights.  \textit{FS} means directly training Stage-2 \textit{from-scratch} without training the Stage-1 model.}
\begin{center}
\resizebox{0.85\columnwidth}{!}{
\begin{tabular}{lcccccc}
\hline
 & \multicolumn{3}{c}{\textbf{Labels $\rightarrow$ Photo}}
 & \multicolumn{3}{c}{\textbf{Photo $\rightarrow$ Labels}} \\
\bf{Method} &  \bf{Pixel acc.} & \bf{Class acc.} &  \bf{Class IoU} &  \bf{Pixel acc.} & \bf{Class acc.} &  \bf{Class IoU} \\
\hline  
CycleGAN~\cite{zhu2017unpaired}  & 0.52 & 0.17 & 0.11 & 0.58 & 0.22 & 0.16 \\
Contrast-GAN~\cite{liang2017generative} & 0.58 & \bf0.21 & \bf0.16 &  0.61 & 0.23 & 0.18 \\
\proposed Stage-1 128 & 0.46 & 0.19 & 0.12 & 0.71 & 0.24 & 0.20 \\
\proposed Stage-1 256 & 0.57 & 0.15 & 0.11 & 0.63 & 0.18 & 0.14 \\
\proposed Stage-2 256-256 & 0.52 & 0.15 & 0.11 & 0.64 & 0.18 & 0.14 \\
\proposed Stage-2 128-256 \textit{FS} & 0.59 & 0.15 & 0.10 & 0.36 & 0.10 & 0.05 \\
\proposed Stage-2 128-256 \textit{FT} & 0.61 & 0.18 & 0.13 & 0.62 & 0.19 & 0.13 \\
\proposed Stage-2 128-256 & \bf0.64 & 0.20 & \bf0.16 & \bf0.72 & \bf0.25 & \bf0.20 \\
\hline
Pix2Pix~\cite{isola2016image} & 0.71 & 0.25 & 0.18 & 0.85 & 0.40 & 0.32 \\
\hline
\end{tabular}
}
\end{center}
\label{tab:city}
\end{table}

\noindent\textbf{Cityscapes \textit{Labels $\leftrightarrow$ Photo.}}
Table \ref{tab:city} shows the comparison of our proposed method \proposed and its variants with state-of-the-art methods in the Cityscapes \textit{Labels $\leftrightarrow$ Photo} tasks. 
The same unsupervised settings are adopted by all methods except Pix2Pix, which is trained under a supervised setting.

On the FCN Scores, our proposed \proposed Stage-2 128-256 outperforms the state-of-the-art approaches considering the pixel accuracy, while being competitive considering the class accuracy and the class IoU. 
On the Segmentation Scores, \proposed Stage-2 128-256 outperforms state-of-the-art approaches in all metrics.
Comparing SCAN Stage-1 256 with CycleGAN, our modified network yields improved results, which, however, still perform inferiorly to SCAN Stage-2 128-256.
Also, we can find that \proposed Stage-2 128-256 achieves a much closer performance to the supervised approach Pix2Pix\cite{isola2016image} than others.

We also compare our \proposed Stage-$2$ 128-256 with different variants of SCAN. 
Comparing \proposed Stage-$2$ 128-256 with \proposed Stage-$1$ approaches, we can find a substantial improvement on the FCN Scores, which indicates that adding the Stage-$2$ refinement helps to improve the realism of the output images.
On the Segmentation Score, comparison of the \proposed Stage-$1$ 128 and \proposed Stage-$1$ 256 shows that learning from low-resolution yields better performance.  Comparison between the \proposed Stage-$2$ 128-256 and \proposed Stage-$1$ 128 shows that adding Stage-2 can further improve from the Stage-1 results.
To experimentally prove that the performance gain does not come from merely adding model capacity, we conducted a \proposed Stage-2 256-256 experiments, which perform inferiorly to the SCAN Stage-2 128-256. 

To further analyze various experimental settings, we also conducted our \proposed Stage-2 128-256 in two additional settings, including \textit{leaning two stages from-scratch} and \textit{fine-tuning Stage-1}.  We  add supervision signals to both stages for these two settings.
Learning two stages from scratch shows poor performance in both tasks, which indicates joint training two stages together does not guarantee performance gain. The reason for this may lie in directly training a high-capacity generator is difficult.
Also, fine-tuning Stage-1 does not resolve this problem and has smaller improvement compared with fixing weights of Stage-1.

To examine the effectiveness of the proposed fusion block, we compare it with several variants: 
1) \textit{Learned Pixel Weight} (LPW), which is our proposed fusion block; 
2) \textit{Uniform Weight} (UW), in which the two stages are fused with the same weight at different pixel locations $\mathbf{\hat{y}_1} (1-w) + \mathbf{\hat{y}_2} w$, and during training $w$ gradually increases from 0 to 1; 
3)  \textit{Learned Uniform Weight} (LUW), which is similar to \textit{UW}, but $w$ is a learnable parameter instead; 
4) \textit{Residual Fusion} (RF), which uses a simple residual fusion $\mathbf{\hat{y}_1} + \mathbf{\hat{y}_2}$.
The results are illustrated in Table~\ref{tab:add}. It can be observed that our proposed LPW fusion yields the best performance among all alternatives, which indicates that the LPW approach can learn better fusion of the outputs from two stages than approaches with uniform weights.

\begin{table}[t]
\small
\tabcolsep=0.11cm
\caption{FCN Scores and Segmentation Scores of several variants of the fusion block on the Cityscapes dataset.} 
\begin{center}
\resizebox{0.8\columnwidth}{!}{
\begin{tabular}{lcccccc}
\hline
 & \multicolumn{3}{c}{\textbf{Labels $\rightarrow$ Photo}}
 & \multicolumn{3}{c}{\textbf{Photo $\rightarrow$ Labels}} \\
\bf{Method} &  \bf{Pixel acc.} & \bf{Class acc.} &  \bf{Class IoU} &  \bf{Pixel acc.} & \bf{Class acc.} &  \bf{Class IoU} \\
\hline  
CycleGAN & 0.52 & 0.17 & 0.11 & 0.58 & 0.22 & 0.16 \\ 
\proposed 128-256 LPW  & \bf0.64 & \bf0.20 & \bf0.16 & \bf0.72 & \bf0.25 & \bf0.20 \\
\proposed 128-256 UW   & 0.59 & 0.19 & 0.14 & 0.66 & 0.22 & 0.17 \\ 
\proposed 128-256 LUW & 0.59 & 0.18 & 0.12 & 0.70 & 0.24 & 0.19 \\
\proposed 128-256 RF   & 0.60 & 0.19 & 0.13 & 0.68 & 0.23 & 0.18 \\
\hline
\end{tabular}
}
\end{center}
\label{tab:add}
\end{table}

In Figure \ref{fig:comp-city}, we visually compare our results with those of the CycleGAN and the Pix2Pix. 
In the \textit{Labels $\rightarrow$ Photo} task, \proposed generates more realistic and vivid photos compared to the CycleGAN. Also, the details in our results appear closer to those of the supervised approach Pix2Pix. 
In the \textit{Photo $\rightarrow$ Labels} task, while \proposed can generate more accurate semantic layouts that are closer to the ground truth, the results of the CycleGAN suffer from distortion and blur.

\begin{figure*}[t]
\begin{center}
   \includegraphics[trim={0 1.5cm 0 0},clip,width=0.72\linewidth]{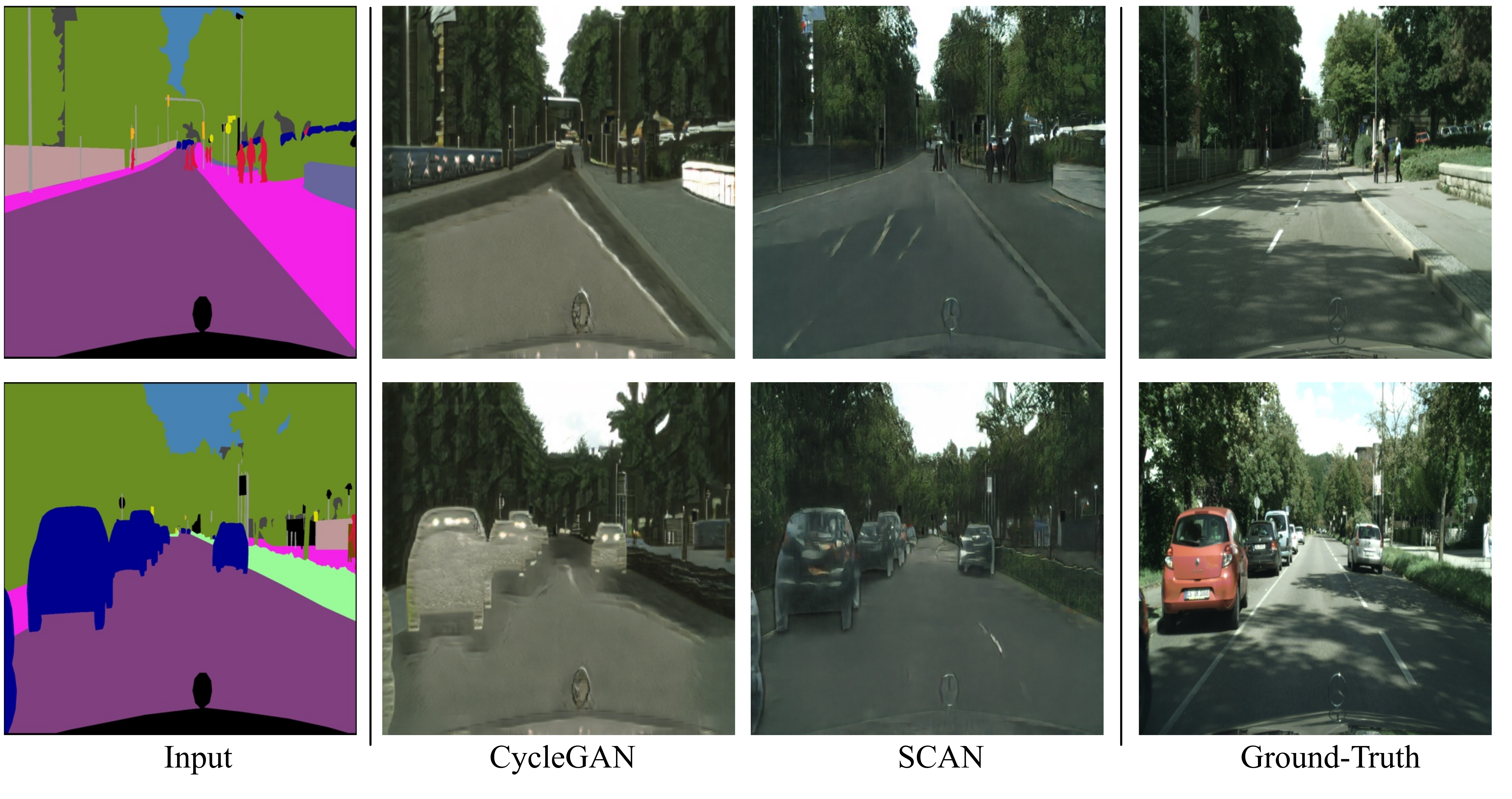}
\end{center}
   \caption{Translation results in the \textit{Labels $\rightarrow$ Photo} task on the Cityscapes dataset of $512 \times 512$ resolution. Our proposed \proposed produces realistic images that even look at a glance like the ground-truths. Zoom in for best view.}
\label{fig:512}
\end{figure*}

\smallskip
\noindent\textbf{High-Resolution Cityscapes \textit{Labels $\rightarrow$ Photo}.} \label{sec:highres}
The CycleGAN only considers images in 256$\times$256 resolution, and results of training CycleGAN directly in 512$\times$512 resolution are not satisfactory, as shown in Figure \ref{fig:1} and Figure \ref{fig:512}.

By iteratively decomposing the Stage-$2$ into a Stage-$2$ and a Stage-$3$, we obtain a three-stage SCAN.
During the translation process, the resolution of the output is growing from $128 \times 128$  to $256 \times 256$ and to $512 \times 512$, as shown in Figure \ref{fig:1}. Figure \ref{fig:512} shows the comparison between our \proposed and the CycleGAN in the high-resolution Cityscapes \textit{Labels $\rightarrow$ Photo} task. We can clearly see that our proposed \proposed generates more realistic photos compared with the results of CycleGAN, and SCAN's outputs are visually closer to the ground truth images. 
The first row shows that our results contain realistic trees with plenty of details, while the CycleGAN only generates repeated patterns. 
For the second row, we can observe that the CycleGAN tends to simply ignore the cars by filling it with a plain grey color, while cars in our results have more details.

Also, we run a user preference study comparing \proposed with the CycleGAN with the setting described in Section~\ref{sec:userstudy}. 
As a result, 74.9\% of the queries prefer our \proposedd's results,  10.9\% prefer the CycleGAN's results, and 14.9\% suggest that the two methods are equal. 
This result shows that our \proposed can generate overall more realistic translation results against the CycleGAN in the high-resolution translation task.

\begin{table}[!t]
\begin{center}
\caption{PSNR and SSIM values in the \textit{Map$\leftrightarrow$Aerial} and \textit{Facades$\leftrightarrow$Labels} tasks.}
\label{tab:mapfacades}
\resizebox{0.75\columnwidth}{!}{
\begin{tabular}{lcccccccc}
\hline
 & \multicolumn{2}{c}{\textbf{Aerial $\rightarrow$ Map}} 
 & \multicolumn{2}{c}{\textbf{Map $\rightarrow$ Aerial}} 
 & \multicolumn{2}{c}{\textbf{Facades $\rightarrow$ Labels}}
 & \multicolumn{2}{c}{\textbf{Labels $\rightarrow$ Facades}} \\
\textbf{Method} &  \textbf{PSNR} &  \textbf{SSIM} &  \textbf{PSNR} & \textbf{SSIM} & \textbf{PSNR} &  \textbf{SSIM} &  \textbf{PSNR} & \textbf{SSIM}  \\
\hline
CycleGAN\cite{zhu2017unpaired} & 21.59 & 0.50 & 12.67 & 0.06 & 6.68 & 0.08 & 7.61 & 0.11 \\
\proposed  & \bf25.15 & \bf0.67 & \bf14.93 & \bf0.23 & \bf8.28 & \bf0.29 & \bf10.67 & \bf0.17\\
\hline
\end{tabular}
}
\end{center}
\end{table}

\begin{figure}[t]
\begin{center}
   \includegraphics[width=0.77\linewidth]{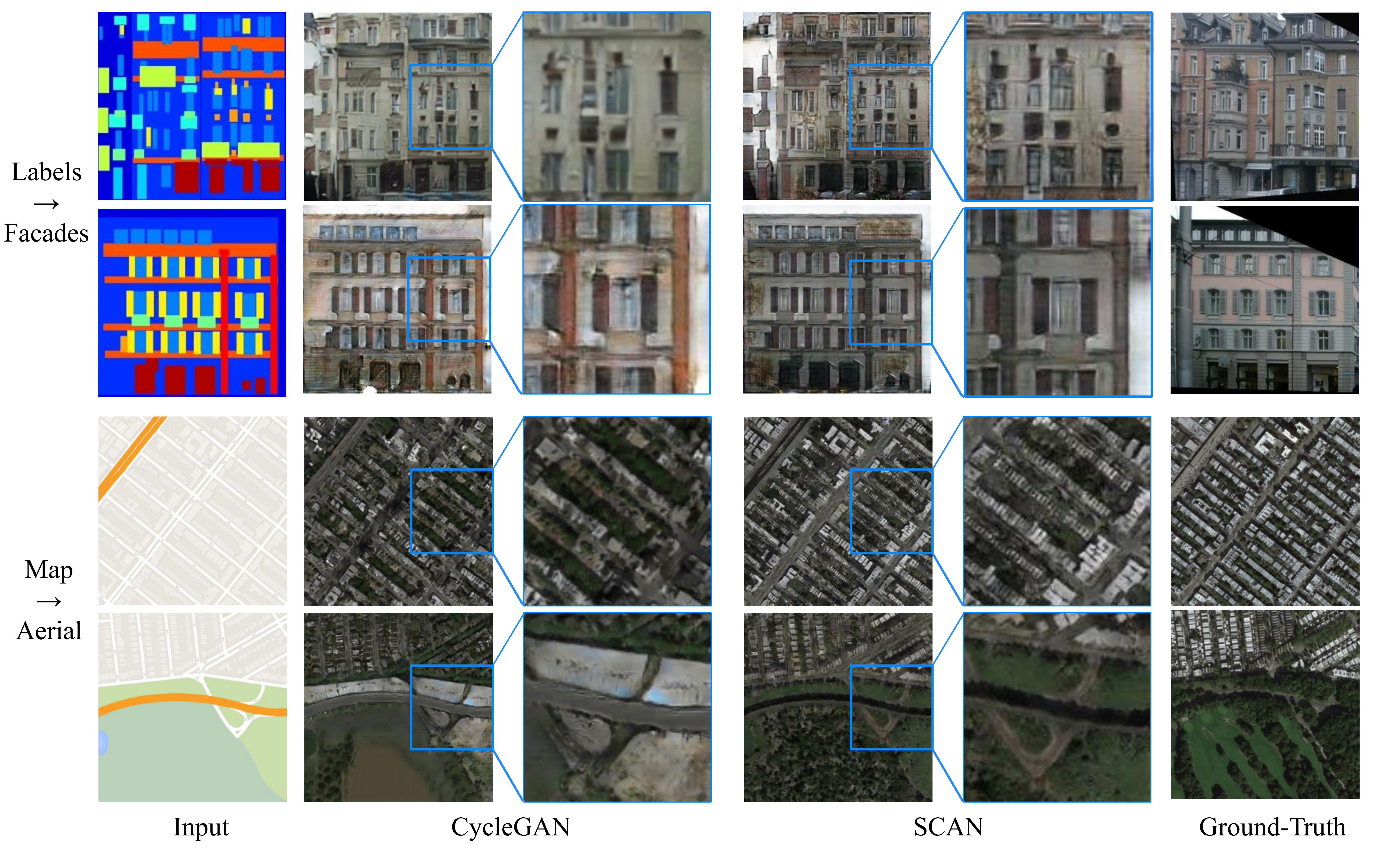}
\end{center}
   \caption{Translation results in the Labels$\rightarrow$Facades task and the Aerial$\rightarrow$Map task. Results of our proposed SCAN show finer details in both the tasks comparing with CycleGAN's results.}
\label{fig:mapfacade}
\end{figure}

\begin{figure}[t]
\begin{center}
   \includegraphics[width=\linewidth]{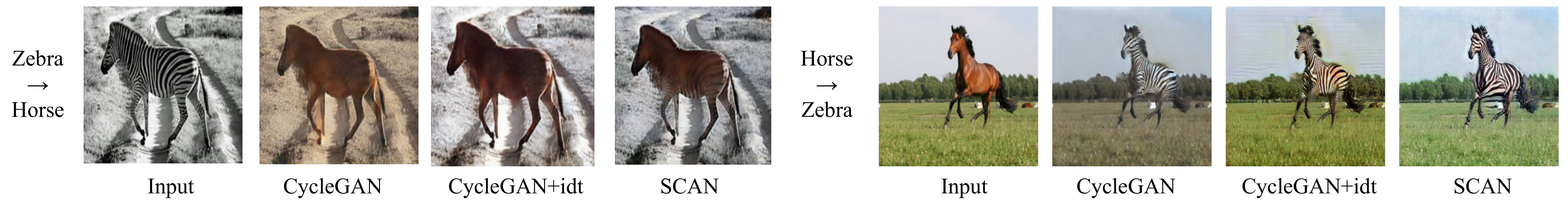}
\end{center}
   \caption{
   Translation results in the Horse$\leftrightarrow$Zebra tasks. CycleGAN changes both desired objects and backgrounds. Adding an identity loss can fix this issue, but tends to be blurry compared with those from SCAN, which never uses the identity loss.
   }
\label{fig:appleorange}
\end{figure}

\smallskip
\noindent\textbf{\textit{Map$\leftrightarrow$Aerial} and \textit{Facades$\leftrightarrow$Labels}.}
Table \ref{tab:mapfacades} reports the performances regarding the PSNR/SSIM metrics. 
We can see that our methods outperform the CycleGAN in both metrics, which indicates that our translation results are more similar to ground truth in terms of colors and structures.

Figure \ref{fig:mapfacade} shows some of the sample results in the Aerial$\rightarrow$Map task and the Labels$\rightarrow$Facades task. 
We can observe that our results contain finer details while the CycleGAN results tend to be blurry.  

\smallskip
\noindent\textbf{\textit{Horse$\leftrightarrow$Zebra}.}
Figure \ref{fig:appleorange} compares the results of SCAN against those of the CycleGAN in the Horse$\leftrightarrow$Zebra task.
We can observe that both \proposed and the CycleGAN successfully translate the input images to the other domain. 
As the Figure \ref{fig:appleorange} shows, the CycleGAN changes not only the desired objects in input images but also the backgrounds of the images.
Adding the identity loss~\cite{zhu2017unpaired} can fix this problem, but the results still tend to be blurry compared with those from our proposed SCAN.
A user preference study on Horse$\rightarrow$Zebra translation is performed with the setting described in Section~\ref{sec:userstudy}. 
As a result, 76.3\% of the subjects prefer our SCAN's results against CycleGAN's, while 68.9\% prefer SCAN's results against CycleGAN+idt's.

\subsection{Visualization of Fusion Weight Distributions}

\begin{figure}[t]
\begin{center}
   \includegraphics[trim={0 2.5cm 0 -1cm},clip,width=0.7\linewidth]{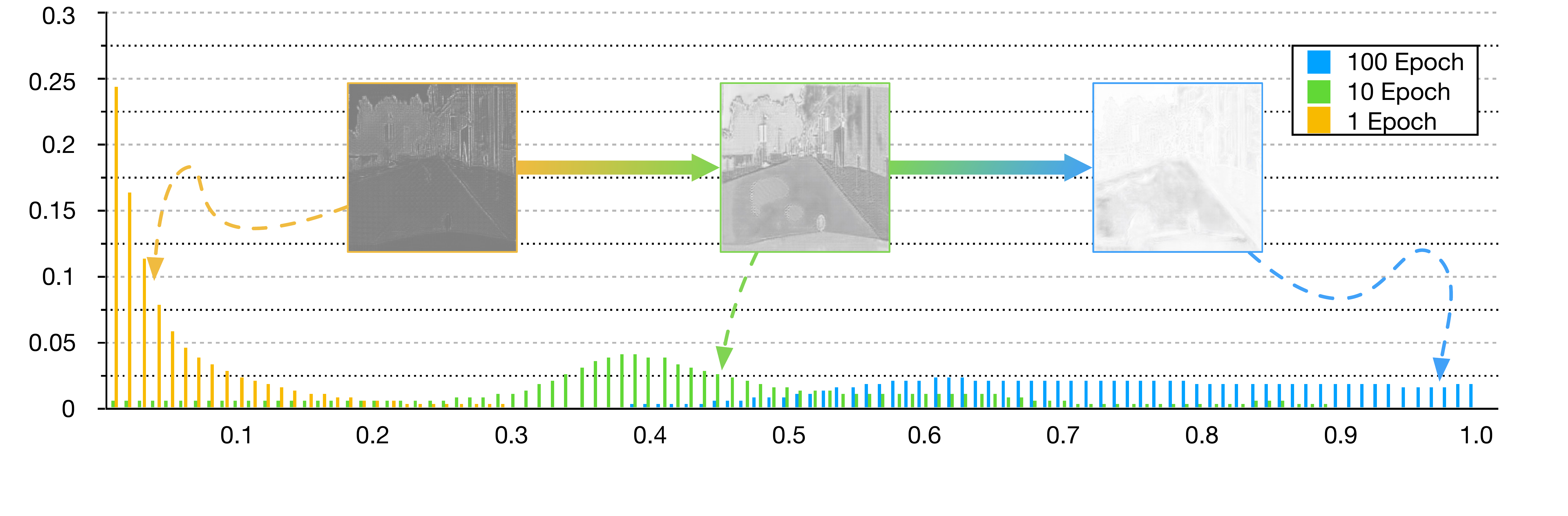}
\end{center}
   \caption{
   \color{black}
   Distributions of fusion weights over all pixels in different epochs. Each distribution is an average result over 1000 sample images from the Cityscapes dataset. 
   Dashed arrows indicate average weights of fusion maps.
   }
\label{fig:alphaevol}
\end{figure}

To illustrate the role of the adaptive fusion block, 
we visualize the three average distributions of fusion weights ($\bm{\alpha}_{x}$ in Equation \ref{equ:alpha}) over 1000 samples from Cityscapes dataset in epoch 1, 10, and 100, as shown in Figure 
\ref{fig:alphaevol}. 
We observed that the distribution of the fusion weights gradually shifts from left to right.
It indicates a consistent increase of the weight values in the fusion maps, which implies more and more details of the second stage are bought to the final output.

\subsection{Ablation Study} \label{sec:ablation}

In Section \ref{sec:comp}, we report the evaluation results of \proposed and its variants, here we further explore SCAN by removing modules from it: 
\begin{itemize}
\item \proposed w/o \textit{Skip} Connection: remove the skip connection from the input to the translation network in the Stage-$2$ model 
, denoted by \textit{\proposed w/o Skip}.
\item \proposed w/o Adaptive \textit{Fusion} Block: remove the final adaptive fusion block in the Stage-$2$ model 
, denoted by \textit{\proposed w/o Fusion}.
\item \proposed w/o \textit{Skip} Connection and Adaptive \textit{Fusion} Block: remove both the skip connection from the input to the translation network and the adaptive fusion block in the Stage-$2$ model 
, denoted by \textit{SCAN w/o Skip,} \textit{Fusion}.
\end{itemize}

\begin{table}[t]
\caption{FCN Scores in the Cityscapes dataset for ablation study, evaluated on the \textit{Labels $\rightarrow$ Photo} task with different variants of the proposed \proposedd.}
\begin{center}
\resizebox{0.72\columnwidth}{!}{
\begin{tabular}{lccc}
\hline
\bf{Method} &  \bf{Pixel acc.} & \bf{Class acc.} &  \bf{Class IoU} \\
\hline  
\proposedsub 128 & 0.457 & 0.188 & 0.124 \\
\proposed Stage-2 128-256 w/o Skip,Fusion & 0.513 & 0.186 & 0.125 \\
\proposed Stage-2 128-256 w/o Skip & 0.593 & 0.184 & 0.136 \\
\proposed Stage-2 128-256 w/o Fusion & 0.613 & 0.194 & 0.137 \\
\proposed Stage-2 128-256 & \bf0.637 & \bf0.201 & \bf0.157 \\
\hline
\end{tabular}
}
\end{center}
\label{tab:fcn-alphaskip}
\end{table}

Table \ref{tab:fcn-alphaskip} shows the results of the ablation study, in which we can observe that removing either the adaptive fusion block or the skip connection downgrades the performance.
With both of the components removed, the stacked networks obtain marginal performance gain compared with Stage-$1$.
Note that the fusion block only consists of three convolution layers, which have a relatively small size compared to the whole network.
Refer to Table \ref{tab:city}, in SCAN Stage-2 256-256 experiment, we double the network parameters compared to SCAN Stage-1 256, resulting in no improvement in the Label $\rightarrow$ Photo task.
Thus, the improvement of the fusion block does not simply come from the added capacity.

Therefore, we can conclude that using our proposed SCAN structure, which consists of the skip connection and the adaptive fusion block, is critical for improving the overall translation performance.

\section{Conclusions}

In this paper, we proposed a novel approach to tackle the unsupervised image-to-image translation problem exploiting a stacked network structure with cycle-consistency, namely SCAN.
The proposed \proposed decomposes a complex image translation process into a coarse translation step and multiple refining steps, and then applies the cycle-consistency to learn the target translation from unpaired image data.
Extensive experiments on multiple datasets demonstrate that our proposed \proposed outperforms the existing methods in quantitative metrics and generates more visually pleasant translation results with finer details compared to the existing methods.


\end{document}